\documentclass[tikz,10pt]{elsarticle} % Use the 'preprint' option for a one-column layout
\usepackage{tikz}
\makeatletter
\def\ps@pprintTitle{%
  \let\@oddhead\@empty
  \let\@evenhead\@empty
  \def\@oddfoot{\reset@font\hfil\thepage\hfil}
  \let\@evenfoot\@oddfoot
}
\makeatother
\usetikzlibrary{shapes.geometric, arrows, positioning, chains}
\usepackage[utf8]{inputenc}
\usepackage{algorithm}
\usepackage{algpseudocode}
\usepackage{amsmath}

\usepackage{amsfonts}
\tikzstyle{startstop} = [rectangle, rounded corners, minimum width=3cm, minimum height=1cm,text centered, draw=black, fill=red!30]
\tikzstyle{process} = [rectangle, minimum width=3cm, minimum height=1cm, text centered, draw=black, fill=orange!30]
\tikzstyle{decision} = [diamond, minimum width=3cm, minimum height=1cm, text centered, draw=black, fill=green!30, aspect=2]
\tikzstyle{arrow} = [thick,->,>=stealth]

\usepackage{graphicx}

\usepackage{colortbl}
\usepackage[margin=2.0cm]{geometry}
\usepackage{hyperref} % To make links clickable
\usepackage{booktabs}
\usepackage{array}
\usepackage{longtable}

\definecolor{Highlight}{RGB}{255, 235, 59} % Lemon color as an example

\title{\textbf{IncidentResponseGPT: Generating Traffic Incident Response Plans with Generative Artificial Intelligence}}
\author{Artur Grigorev (\url{Artur.Grigorev@uts.edu.au}), University of Technology Sydney, Australia\\ Dr Khaled Saleh (\url{Khaled.Saleh@ieee.org}), University of Newcastle, Australia\\ Dr Yuming Ou (\url{Yuming.Ou@uts.edu.au}), University of Technology Sydney, Australia\\
 Dr Adriana-Simona Mihaita (\url{Adriana-Simona.Mihaita@uts.edu.au}), University of Technology Sydney, Australia\\}
\date{}

\begin{document}
\begin{abstract}
The proposed IncidentResponseGPT - a novel framework that applies generative artificial intelligence to generate response plans based on incident response guidelines and incoming incident reports. This model allows for synthesis of incident response guidelines and generates incident response plans adapted to specific area, aiming to facilitate the decision-making process for traffic management authorities. This approach aims to accelerate incident resolution times by suggesting various actions (e.g. the use of variable message signs) to minimize the overall impact on the urban traffic network. The system suggests specific actions, including lane closures, announcements and dispatching appropriate emergency resources. We also propose the novel application of the Technique for Order Preference by Similarity to Ideal Solution (TOPSIS) to rank generated response plans based on criteria like impact minimization and resource efficiency and compare solutions based on their proximity to an human-proposed solution.

\end{abstract}

\maketitle

% \end{document}

\section{Introduction}

% Traffic congestion caused by road incidents represents a critical challenge in urban environments, leading to a cascade of detrimental impacts. Increased travel times, elevated pollution levels, and significant economic losses due to delays are just a few of the consequences faced by commuters and cities alike. While existing traffic management centers (TMCs) employ a mix of automated and manual processes \cite{de2017traffic,s16020157}, these systems often struggle to keep pace with the increasing complexity and real-time demands of today's traffic conditions.  

% Addressing this critical need, we introduce IncidentResponseGPT, a novel system designed to revolutionize traffic incident response by leveraging the power of generative artificial intelligence (AI). Unlike traditional approaches, IncidentResponseGPT goes beyond simple automation. It analyzes real-time traffic incident reports and integrates region-specific operational guidelines to rapidly generate adaptable response plans. This empowers traffic management authorities with a proactive and intelligent tool to significantly improve the efficiency and effectiveness of their incident response strategies. This is achieved by incorporating a Generative AI \cite{baidoo2023education} with traffic incident reports and area-specific operational guidelines.

Traffic congestion from road incidents poses significant challenges, including increased travel times, pollution, and economic losses. While Traffic Management Centers (TMCs) utilize a combination of automated and manual processes \cite{de2017traffic,s16020157}, these systems often fail to meet the growing complexity and real-time demands of modern traffic conditions which create the demand for rapid and sophisticated decisions.

To address this, we introduce IncidentResponseGPT, a generative AI-based system designed to enhance traffic incident response. By analyzing real-time traffic reports and integrating region-specific guidelines, it generates dynamic response plans, offering a proactive tool to improve incident management efficiency.

While promising, the integration of AI into current traffic management workflows presents the following inherent challenges:
\begin{enumerate}
    \item \textbf{Complexity of Urban Traffic Network}: Urban traffic networks are extraordinarily complex systems with various intersecting roads, bridges, and tunnels. This complexity makes it difficult to predict and manage traffic during incidents effectively.
    
    \item \textbf{Variety of Traffic Incidents}: The nature and severity of traffic incidents can greatly vary, including minor accidents, major crashes, natural disasters, road works, etc. Each type requires a different and suitable response strategy.
    
    \item \textbf{Real-Time Decision Making}: Traffic management authorities must make quick decisions. Any delay could lead to a change in a situation making timely response plans necessary. Any change or incoming information also necessitates the re-evaluation and adaptation of initially proposed plan.
    
    \item \textbf{Alignment with Local Laws and Regulations}: Different regions have different response guidelines and roadway regulations. The decision-maker must consider specific regulations when managing any road situation. 
\end{enumerate}

The novelty of the IncidentResponseGPT framework is in its unique application of generative AI within the domain of traffic management. Our system can synthesize incident response guidelines into a structure tabular form and propose incident response plans that are adapted to the specific circumstances of each traffic incident. We also introduce a method to quantify the generated incident plans (via binary array representation). This approach holds the potential to not only accelerate incident resolution times but also to minimize the overall impact of incidents (or other traffic events involving decision-making) on the entire urban traffic network.

% . These include effectively representing the complexity of urban road networks, accounting for the wide variety of potential incident types, ensuring real-time decision-making capabilities, and adhering to local traffic laws and regulations:

% overall, IncidentResponseGPT presents a novel approach with the capability to revolutionize traffic incident management in urban environments. By using the power of generative AI, our proposed system aims to overcome the limitations of traditional approaches by offering rapid, adaptable, and informed response plans. The potential benefits are significant, promising to reduce congestion, minimize economic losses, and enhance overall safety for all stakeholders within the urban traffic ecosystem.

\section{Related Works}

Large Language Models (LLM) have shown promise in disaster management, demonstrating their ability to generate rapid response plans and disseminate information \cite{xue2023, goecks2023}. However, these models often struggle with real-time data integration and may lack the domain-specific knowledge required for accurate traffic incident management. Similarly, existing AI-based traffic management systems often rely on techniques like Case-Based Reasoning (CBR) or Bayesian Theory \cite{ma2014automated, ravish2021intelligent}. While these approaches have shown some success, they often struggle with the scalability and adaptability required for real-time incident response. CBR systems are heavily reliant on the quality and quantity of historical cases, which can limit their effectiveness in novel situations. Bayesian methods, while powerful, can become computationally expensive in real-time scenarios, especially with large datasets \cite{ma2014automated}. 

Case-Based Reasoning (CBR) and Bayesian Theory have been utilized to automatically generate traffic incident response plans [\cite{ma2014automated}]. However, the effectiveness of CBR depends on the quality and quantity of historical cases in the database, as limited or biased data can lead to suboptimal decision-making. Moreover, calculating posterior probabilities using Bayesian Theory in real time can be resource-intensive, particularly with complex models and large data sets.
% Current studies provides substantial insights into the utilization of large language models and various methods of artificial intelligence (AI) to enhance disaster response, incident prediction, and transportation management. Given the increasing frequency and impact of natural disasters and the increase in complexity of traffic scenarios across the globe, it becomes imperative to address the efficiency of responses and the methodologies of implementing AI-driven response planning within these dynamic environments. 

The OpenAI's ChatGPT model previously has been explored for its capabilities in natural disaster management and demonstrates the applicability of LLM models in accelerating the disaster management process through stages such as pre-disaster, during-disaster, and post-disaster scenarios \cite{xue2023}. These models prove effective in generating rapid response plans and disseminating crucial information during the event, with inherent limitations concerning real-time data processing and professional accuracy. The model addresses the limitations in real-time data handling and expert-level accuracy in complex disaster management scenarios. Also, a recently proposed DisasterResponseGPT framework \cite{goecks2023} relies on LLMs to generate rapid action plans in disaster situations with limitations in handling visual and real-time data inputs. The discussion on using simulation models like Visual Interactive System for Transport Algorithms (VISTA) for simulating traffic conditions and test incident management strategies further underscores the benefits of dynamic modeling in responding to varying traffic conditions effectively \cite{wirtz2005using}. These simulations help in preplanning which, as their findings suggest, is essential for reducing response times and managing congestion effectively during traffic incidents.

% Various works highlight trust issues with AI-generated content during disasters, promoting transparency as a pivotal factor in enhancing public trust and plan reliability \cite{tao2023}.

Through integrating Geographic Information System (GIS) capabilities with AI \cite{huang2007}, the efficiency of incident responses can be markedly improved by optimizing resource allocation and managing traffic disruptions in real-time scenarios. AI models are capable of facilitating quick and effective decision-making processes by processing and analyzing large volumes of data swiftly \cite{huang2007}. Works that tackle the integration of AI in traffic systems further affirm the potential of LLMs in transforming urban traffic management \cite{zhang2024}. The integration of real-time data, the need for continuous model training, and the development of robust frameworks to handle complex emergency situations are recurrent themes across studies. As indicated by safety analytics in transport and construction sectors, there remains a crucial gap in applying AI to dynamically interpret complex environmental data sets and interact seamlessly across diverse operational platforms. Real-time traffic incident management demands rapid assessment and response, which can be significantly optimized using Generative AI. There is a critical need for integrating real-time data with AI systems to manage incidents effectively on complex traffic networks \cite{huang2007integration}.

The integration of data-driven models in Emergency Response Management systems was previously extensively reviewed \cite{mukhopadhyay2023emergency} and highlighted the disconnect between traditional static models and the dynamic requirements of real-world emergency management. The authors proposed a transition towards more adaptive models, capable of real-time adjustments in response strategies based on live data inputs. Key challenges identified include the ethical implications which remain paramount as these technologies become more relevant in critical areas affecting public safety of predictive policing and the limitations of current technology in accurately modeling complex urban behaviors. Current study aims to address AI capabilities with a consideration of ethical implications, striving for transparency and reliability of AI-driven methodology in traffic incident management. The application of dynamic resource allocation has been demonstrated in a similar case of response planning: crime response, with strategies like predictive analytics and hotspot policing gaining prominence. The continuous-time Markov decision process (CTMDP) uses real-time crime data to optimize police patrol allocations, significantly improving response times in urban scenarios by minimizing the duration between crime occurrence and police arrival.

% The potential of Generative AI to enhance traffic incident management significantly revolves around its ability to process and model the incident data. 

   %The introduction of IncidentResponseGPT addresses these challenges by combining generative AI with incident report data. % with the consideration of the importance of real-time adaptable plans in managing urban disasters \cite{goecks2023}. 

% The evolution of AI technologies offers promising enhancements in managing urban traffic incidents.
% IncidentResponseGPT demonstrates the application of incorporating accident report data and generative AI to formulate incident response strategies. 
% The proposed framework can be adapted to produce responses based on continuously updated traffic data, operator input and incident progression \cite{mukhopadhyay2023emergency}, due to textual nature of Generative AI and both historical and real-time accident reports.

Despite recent advancements, several challenges persist in the use of AI for incident management. Recent study on data-driven models in emergency response management identified the need for adaptive models capable of real-time adjustments based on live data inputs \cite{mukhopadhyay2023emergency}.  Trust and transparency in AI-generated content during emergencies are crucial \cite{tao2023}, emphasizing the need for transparent decision-making processes, which IncidentResponseGPT addresses through its chain-of-thought visualization and validation process.

% The primary stakeholders in AI-driven traffic incident management solutions include (1) traffic management authorities and municipal bodies responsible for traffic control and road infrastructure, and (2) emergency response teams, such as the police, fire services, and medical responders, who play a crucial role in managing incidents.

Ravish and Swamy's review discusses the integration of machine learning and computational intelligence in managing traffic congestion, highlighting significant advancements in Intelligent Transportation Systems (ITS). They stress the importance of real-time data processing and accident detection mechanisms for improving traffic flow and safety [\cite{ravish2021intelligent}]. The review identifies challenges such as the increasing number of vehicles and insufficient infrastructure expansion. Although machine learning and computational intelligence offer promising solutions, real-time data processing, system complexity, adaptability, and scalability, as well as accident detection, remain critical concerns for the advancement and expansion of traffic systems.

 The use of generative AI addresses the limitations of static area-specific models \cite{mukhopadhyay2023emergency} through the flexibility of using information represented as a text. It allows to promptly change guidelines for incident response or even try guidelines from other countries or areas to compare their efficiency. Also, it allows for adaptation of insights from safety guidelines and analytics from other sectors, like transportation construction area \cite{smetana2024highway}, to the domain of traffic incident response (e.g. the description of a highway architecture can help to derive more efficient plans with consideration of its operational and structural specifics). AI models demonstrate capabilities in performing complex text analysis, including the examination of accident reports to enhance the prediction of various aspects of traffic incidents \cite{zheng2023chatgpt}. However, the potential of Generative AI in generating immediate incident response plans remains largely unexplored.

 Overall, this study underscores the capacity of LLMs to process complex information from the area of traffic management and generate actionable safety recommendations, illustrating the broader applicability of AI in managing hazards.

\section{Methodology and Data flow}

\begin{figure*}[htbp]
  \centering
  \includegraphics[width=0.8\textwidth]{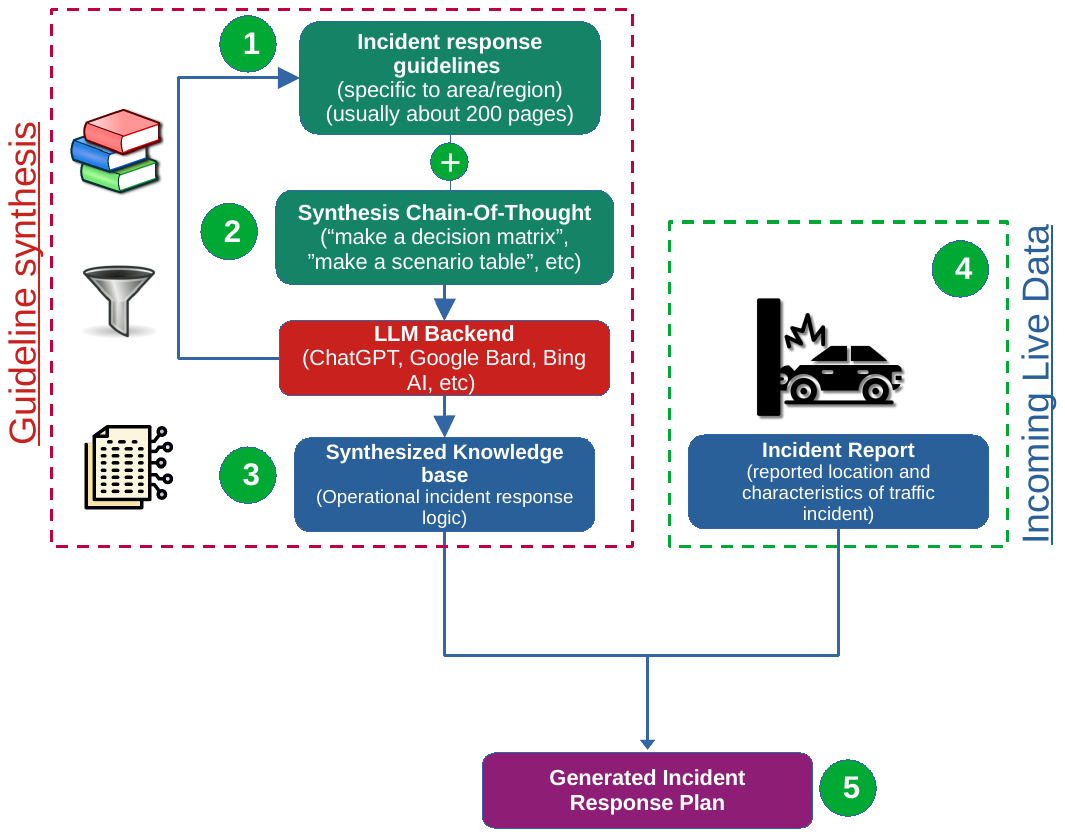}
  \caption{TrafficIncidentResponseGPT: data flow diagram}
  \label{fig:example}
\end{figure*}

We use various Generative AI models within the framework that can generate the response plans based on accident report and incident response guidelines (see Figure \ref{fig:example}). The process includes two major steps: a) Incident response guidelines get processed into synthesized form using the \textbf{Guideline synthesis} procedure utilizing one of LLM models (the model with the highest performance and largest context window is preferable at this step since the scale and complexity of guidelines), b) Synthesized representation of guidelines gets combined with an incident report by concatenation with the prompt in order to obtain the incident response plan.

The \textbf{Guidelines} - "Traffic Incident Management Handbook" \cite{FHWA2010TIMHandbook} as incident response guidelines. These guidelines were prepared for the Federal Highway Administration in January 2010, serve as a comprehensive guide for agencies involved in managing traffic incidents on public roadways. It targets two key audiences: 1) Managers responsible for developing and improving incident management programs, 2) Field Practitioners tasked with the day-to-day execution of these programs. The handbook addresses a wide range of topics related to incident management, including: 1) definitions for incident management, outlines its significance, and identifies the key agencies involved, 2) organization, planning, and design of an Incident Management, 3) provides a step-by-step process for establishing or enhancing programs, emphasizing the importance of collaboration, data analysis, goal setting, and evaluation, 4) outlines operational and technical approaches to improving the incident management process with the mention of specific techniques and strategies for each stage of incident management, such as detection, verification, motorist information, response, site management, traffic management, and clearance, 5) combines process-oriented guidance for program development with a toolbox of strategies and techniques applicable to real-world incident response. Within our framework, these guidelines pre-processed using Generative AI models to extract key instructions related to incident severity estimation, responder safety, traffic control, and information dissemination. The page length of these guidelines is 116 pages. These extracted guidelines are then converted into a structured tabular format. These steps are necessary to allow the AI model to utilize instructions effectively with the consideration for inherent limitation for context window of Generative AI models.

The \textbf{Accident reports} - \textbf{US-Accidents dataset} is a comprehensive countrywide traffic accident dataset for the United States, excluding Alaska and Hawaii, that includes about 1.5 million traffic accident records collected continuously since February 2016 from multiple data providers. The data captures real-time traffic events through several APIs sourced from different entities such as US and state departments of transportation, law enforcement agencies, and traffic cameras.

The dataset includes a number of attributes for each accident record:
\begin{itemize}
    \item \textbf{Location Details:} Latitude and longitude, city, county, state, and zip code.
    \item \textbf{Time Details:} Accident start and end time, timezone, and timestamps for weather observations.
    \item \textbf{Environment:} Weather conditions such as temperature, wind chill, humidity, visibility, and precipitation.
    \item \textbf{Accident Details:} Severity, description, and distance affected.
    \item \textbf{Nearby Features:} Presence of traffic signals, crossings, junctions, and other points of interest.
    \item \textbf{Daylight Conditions:} Based on sunrise, sunset, and various levels of twilight.
\end{itemize}

In a real-world implementation, Generative AI models have a property of high adaptability to available data sources due to its inherent ability to process any information represented as text. This could potentially allow the system to continuously adapt its response plan based on real-time updates from the field, traffic sensors, and other data sources.

The core of the IncidentResponseGPT system is powered by a generative language model like OpenAI ChatGPT / Google Gemini. Based on text-processing capabilities of Generative AI models, the response plans can not only be based on incoming incident data adhering to regional-specific incident response guidelines, but also real-time feedback from an operator and the current and progressive state of the traffic incident. This adaptability enhances the flexibility and situational appropriateness of the response plans, leading to improved traffic safety, reduction of congestion, and ultimately, lesser economic and environmental impacts of traffic incidents. 

% To facilitate transparency of decision-making process, IncidentResponseGPT provides a Chain-of-thought visualization.  This visualization highlights which instructions from the handbook were prioritized and how they influenced the specific actions within the response plan. For example, the system might show that the decision to prioritize a nearby tow truck was based on the guideline emphasizing the rapid clearance of blocked lanes.

Our system performs analysis and synthesis of traffic incident guidelines into a short actionable form to address inherent limitations of context size experienced with these models. We call this process as \textbf{'Guideline Synthesis'}, the results of which can be ‘plugged-into’ the AI system to generate incident response according to specific guidelines.

\section{Guideline synthesis}
% \begin{centered}

\subsection{S-cycle}

\begin{figure*}[htbp]
  \centering
  \includegraphics[width=0.8\textwidth]{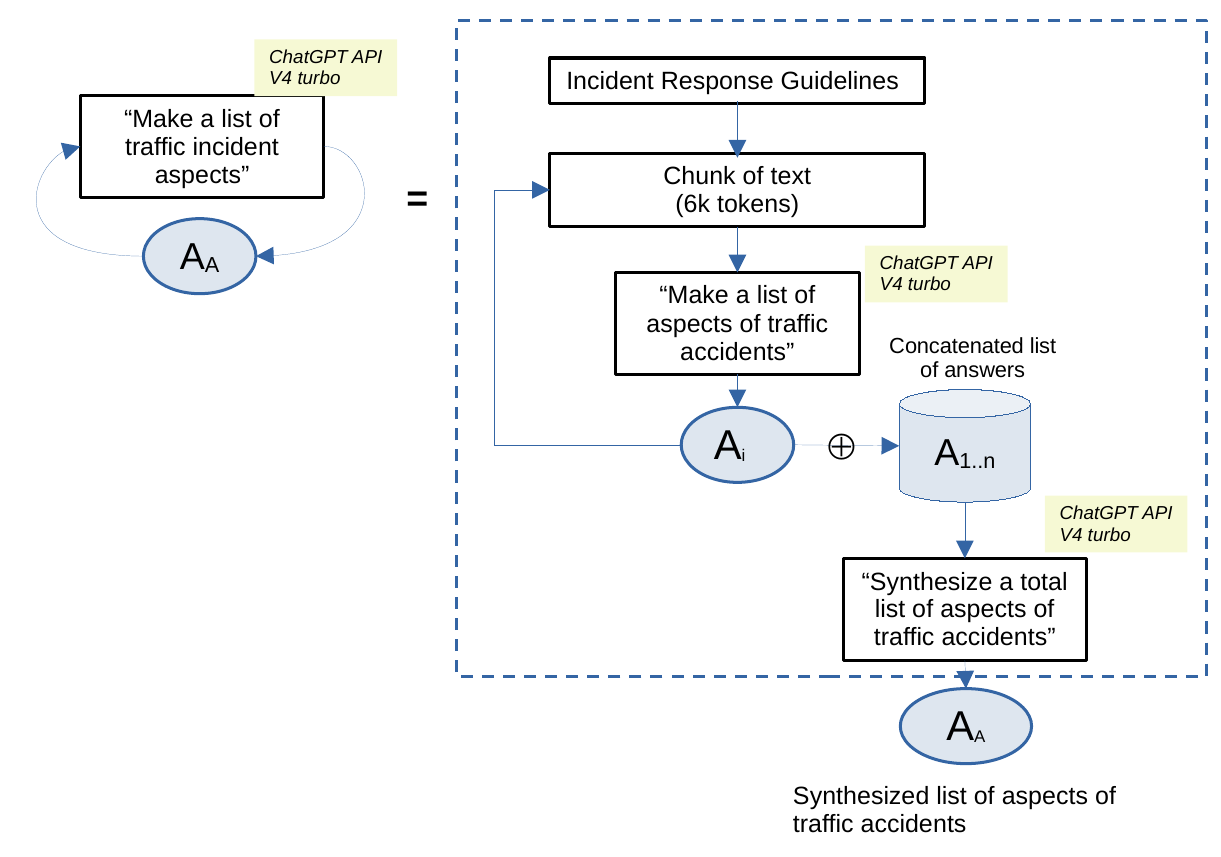}
  \caption{Cyclic processing of text chunks with synthesis
(S-cycle). Example of incident aspects extraction}
  \label{fig:cycle}
\end{figure*}

We introduce a novel chain-of-thought pattern named \textbf{S-cycle} - the core operation of guideline synthesis procedure that consists of multiple processing and synthesis cycles of text chunks. This procedure can be used as a component of a larger chain-of-thought schema and allows to process large amounts of text. S-cycle consists of the following steps (see Figure \ref{fig:cycle}):
\begin{itemize}
\item Text Chunking: The first step involves breaking down the given full text of incident response guidelines (guidelines can be extensive and include up to 300 pages) into manageable segments, or "chunks," to facilitate processing. This step acknowledges the limitations in processing capacity of generative models for large texts and ensures that each segment can be individually summarized. For our current research, the size of these chunks is defined at 6000 tokens to address limited token space of various LLM models.

\item Iterative Processing: The core of the process involves iterative steps where each text chunk is processed individually. This iterative process is initiated with a "For Each Chunk" node, indicating the beginning of a loop that applies to every chunk derived from the original text.

\item Query Application: For each chunk, a query, specifically "summarize as a table," is applied using ChatGPT technology. This step leverages the natural language processing and generative capabilities of ChatGPT to convert textual information into a structured table format, summarizing the key points of each chunk.

\item Result Acquisition: Following the application of the query, the result, which is the ChatGPT-generated table summarizing the chunk, is obtained.

\item Loop Continuation Decision: After processing a chunk, a decision node "End of Text?" checks if there are more chunks to process. If more chunks remain ("no"), the process loops back to continue with the next chunk. This iterative cycle ensures that every chunk of the original text is processed and summarized.

\item Response Combination and Final Synthesis: Once all chunks have been processed and summarized, the resulting tables from each chunk are combined. A final query, "synthesize the table," is applied to this combined data, aiming to integrate the content into a single, coherent table. This step is crucial for synthesizing the fragmented summaries into a unified representation of the original text.

\item Final Output: The process is finished with obtaining the final table, which represents a synthesized summary of the entire original text in a structured table format.

\end{itemize}

% Our research outlines a strategic approach for processing and synthesizing methodologies related to traffic incident response management, structured around four types of queries.

\begin{figure*}[htbp]
  \centering
  \includegraphics[width=0.8\textwidth]{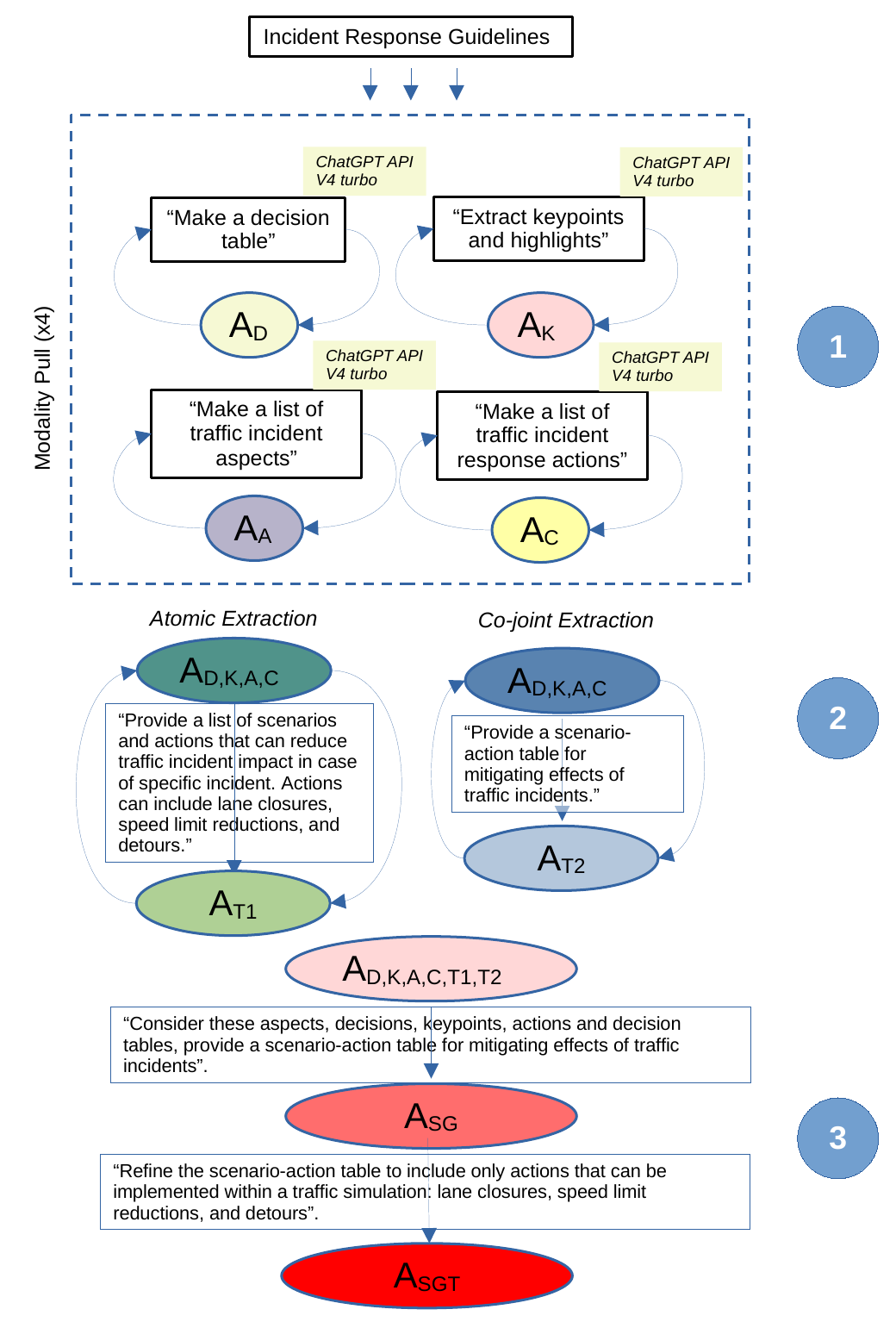}
  \caption{Dataflow for guideline synthesis}
  \label{fig:synthesis}
\end{figure*}

\begin{table}[h]
\centering
\scriptsize
\caption{ChatGPT4 API output: Traffic Incident Mitigation Strategies}
\label{tab:traffic_incident_mitigation}
\begin{tabular}{|p{1cm}|p{2cm}|p{1cm}|p{3cm}|p{4cm}|p{4cm}|}
\hline
\textbf{Scenario ID} & \textbf{Incident Type} & \textbf{Severity} & \textbf{Location} & \textbf{Action} & \textbf{Equipment/Technology Required} \\
\hline
1 & Vehicle Breakdown & Low & Highway/Freeway Lane & - Deploy Incident Response Vehicle\newline - Temporary lane closure\newline - Use VMS to warn drivers & - Service Patrol Vehicle\newline - Variable Message Signs (VMS)\newline - Traffic Cones or Barriers \\
\hline
2 & Minor Two-Car Collision & Moderate & Urban Arterial & - Notify Police \& EMS if needed\newline - Quick clearance policy\newline - Use VMS \& social media & - Police/EMS Vehicles\newline - Quick Clearance Equipment (e.g., tow truck)\newline - Variable Message Signs\newline - Social Media Platforms \\
\hline
3 & Major Multi-Vehicle Crash & High & Highway/Freeway & - Full or partial lane closures\newline - Divert traffic to detour routes\newline - Activate EOC & - Police/EMS/Fire Services\newline - Emergency Operations Center (EOC) Activation\newline - Detour Signage \\
\hline
4 & Hazardous Material Spill & High & Near Urban Area & - Full road closure\newline - Mandatory evacuation if necessary\newline - HazMat Team dispatch & - HazMat Team\newline - Road Closure Signage\newline - Emergency Alert System (EAS) \\
\hline
5 & Overturned Truck & Moderate & Highway On-ramp/Off-ramp & - Partial ramp closure\newline - Speed limit reduction in area\newline - Deploy tow truck and cleanup crew & - Tow Trucks\newline - Cleanup Crew \newline - Speed Limit Signs \\
\hline
6 & Pedestrian Accident & High & Urban Crosswalk & - Full closure of affected lanes\newline - EMS priority dispatch\newline - Investigative procedures & - EMS Vehicles\newline - Police Investigation Unit\newline - Temporary Signage \\
\hline
7 & Wildlife on Road & Low & Rural Road & - Temporary speed limit reduction\newline - Use VMS to warn drivers\newline - Wildlife control dispatch & - Variable Message Signs\newline - Wildlife Control Services \\
\hline
8 & Infrastructure Failure (Bridge) & Very High & Bridge & - Full bridge closure\newline - Long-term detour setup\newline - Structural assessment & - Structural Engineering Team\newline - Permanent Detour Signage\newline - Media Briefing Equipment \\
\hline
9 & Snow/Ice Conditions & Variable & Major Roadways & - Speed limit reduction\newline - Deploy snow plows/salt trucks\newline - Use VMS and radio to inform & - Snow Plows/Salt Trucks\newline - Variable Message Signs\newline - Radio Broadcast System \\
\hline
10 & Fog/Visibility Issues & Variable & Highway/Freeway & - Speed limit reduction\newline - Flashing lights to warn drivers\newline - Use VMS to advise caution & - Variable Message Signs\newline - Highway Flashing Lights \\
\hline
\end{tabular}
\label{tab:guide}
\end{table}

\subsection{Guideline synthesis}
The guideline synthesis procesure is a complicated multi-step approach (see Figure \ref{fig:synthesis}):
\begin{itemize}
\item \textbf{Modality Pull} - the chain-of-thought pattern that we introduce to address the task of text decomposition / information compression. The idea is to perform some form of projection of text into feature/concept space. Inspired by the kernel trick from SVM, which projects data into higher-dimensional spaces to make it linearly separable \cite{scholkopf2000kernel,gordon2004support}, this step allows for decomposing text into structured elements across different modalities. We perform 'projection' from text into following modalities using dedicated S-cycles: a) decision tables b) keypoints and highlights c) incident aspects d) response actions
\item \textbf{Atomic} and \textbf{Co-joint} extraction - also introduced patterns due to our observation that LLM perform extraction of separate facts from text is much more efficient rate than asking to generate scenario-action pairs, which implies reasonong which puts more pressure on models reasoning capabilities and space for error.
\item \textbf{Mass-concatenation} - the results of the S-cycles, which generate reduced and refined information, are concatenated into a final prompt for generating response plans.
\item \textbf{Refinement step} - asking the system overcomplicated prompts or asking to solve multiple tasks at once, by our observation, produces significantly suboptimal results. That's why the generation of scenario-action table consists of two steps - asking for the generation of the table and asking to refine or filter it.

\end{itemize}

\subsection{Derivation of incident response plans}
\begin{figure*}[htbp]
  \centering
  \includegraphics[width=0.8\textwidth]{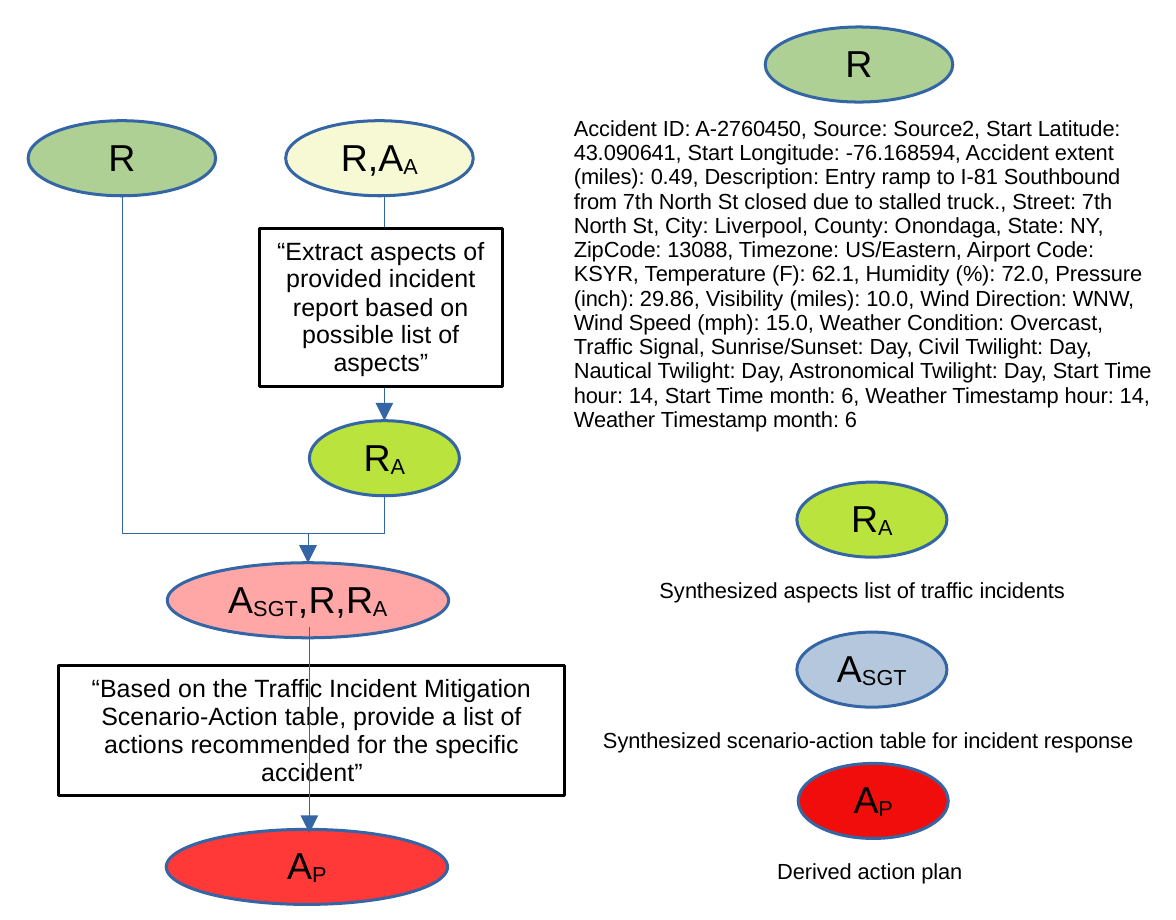}
  \caption{Derivation of actionable response plan}
  \label{fig:plan}
\end{figure*}

Derivation of actionable response plan consists of combination of extracted accident aspects, synthesized guidelines and actual accident report (see Figure \ref{fig:plan}). Then, synthesized guidelines combined with refined accident description to produce a recommendation of actions to reduce the accident impact. The following Table \ref{tab:guide} represents a sample output of Guideline Synthesis procedure. 
% The prompt is applied to convert to a form which is easier to comprehend by LLM due to narrative nature of majority of datasets used for its training. 

Table \ref{tab:example} represents an example traffic accident report which will be used to produce incident response plans using various LLM models.
\begin{table}[h]
\centering
\caption{Accident Information}
\begin{tabular}{|p{14cm}|}
\hline
Accident ID: A-2760450, Source: Source2, Start Latitude: 43.090641, Start Longitude: -76.168594, Accident extent (miles): 0.49, Description: Entry ramp to I-81 Southbound from 7th North St closed due to stalled truck., Street: 7th North St, City: Liverpool, County: Onondaga, State: NY, ZipCode: 13088, Timezone: US/Eastern, Airport Code: KSYR, Temperature (F): 62.1, Humidity (\%): 72.0, Pressure (inch): 29.86, Visibility (miles): 10.0, Wind Direction: WNW, Wind Speed (mph): 15.0, Weather Condition: Overcast, Traffic Signal, Sunrise/Sunset: Day, Civil Twilight: Day, Nautical Twilight: Day, Astronomical Twilight: Day, Start\_Time\_hour: 14, Start\_Time\_month: 6, Weather\_Timestamp\_hour: 14, Weather\_Timestamp\_month: 6\vspace{0.3cm}
\end{tabular}
\label{tab:example}
\end{table}

OpenAI ChatGPT, Microsoft Copilot and Google Gemini were used to generate an incident response plan according to synthesized guideline table and accident description (see Table \ref{table:response_strategies}).

Each AI system has proposed a set of actions, equipment, and additional considerations to mitigate the effects of the incident, based on the provided Traffic Incident Mitigation Strategies (see Tables \ref{tab:incident_response_plan_gpt}-\ref{tab:incident_response_plan_ge}). Each AI system offers a structured response that aligns with the characteristics of the accident, with slight variations in focus and equipment recommendations based on their interpretation of the scenario.

% Table \ref{tab:incident_response_plan_gpt} illustrates the response plan generated by ChatGPT. The plan suggests the deployment of essential safety measures, including the partial closure of the ramp, a temporary speed limit reduction, and the use of a tow truck to remove the stalled truck.

% Table \ref{tab:incident_response_plan_co} outlines the response strategy proposed by Microsoft Copilot. Similar to the ChatGPT plan, it emphasizes the need for ramp closure, speed reduction, and tow truck deployment, while also stressing the importance of prioritizing road user safety and rapid clearance to restore traffic flow.

% Table \ref{tab:incident_response_plan_ge} presents the Google Gemini incident response plan. This plan recommends deploying an incident response vehicle, closing the entry ramp to I-81 Southbound, and using Variable Message Signs (VMS) to warn drivers of the stalled truck and lane closure. It also highlights the low severity of the accident and advises drivers to use alternate routes.

ChatGPT and Microsoft Copilot provided a closely aligned response plan with partial ramp closure, speed limit reduction, and deployment of tow trucks and cleanup crews, closely mirroring the guidelines for a “Moderate” severity incident involving a truck. Google Gemini, while suggesting immediate safety measures, seems slightly less detailed in specifying the sequential steps or acknowledging the cleanup process which might be necessary for such incidents.

Both ChatGPT and Microsoft Copilot demonstrated a higher level of detail in their response plans, matching the granular approach recommended in the provided traffic incident mitigation strategies. Google Gemini focused on broad strokes, which could be interpreted as a more general approach suited for a quick initial response but less detailed for full incident management.

The equipment recommendations from ChatGPT and Microsoft Copilot were closely aligned with synthesized guidelines for handling an incident involving a stalled or overturned truck, demonstrating a strong adherence to the predefined response planning table. Google Gemini’s recommendations were also appropriate but lacked mention of a cleanup crew, which could be integral to the scenario described. 

% ChatGPT and Microsoft Copilot both adhered closely to the provided guidelines, with a comprehensive and detailed approach to incident response, including both immediate safety measures and subsequent actions for traffic management and site clearance. Google Gemini, while adhering to the safety-first principle, could have expanded on the follow-up actions and detailed equipment use, offering a slightly less detailed plan when compared directly with the predefined traffic incident mitigation strategies. 

Google Gemini Flash is offering a slightly less detailed plan, could have expanded on the follow-up actions and detailed equipment use. No model significantly diverged from the framework; however, the differences is in the level of detail and granularity in the action plans and equipment listings, with ChatGPT and Microsoft Copilot providing a more in-depth, scenario-specific response.

% \begin{center}
% \begin{longtable}{|p{14cm}|}
% \hline
% \caption{ChatGPT's Incident Response Plan for Accident ID A-2760450}\\
% \hline
% \endfirsthead
% \multicolumn{1}{c}
% {{\tablename\ \thetable{} -- continued from previous page}} \\
% \hline 
% \endhead
% \hline \multicolumn{1}{r}{{Continued on next page}} \\ 
% \endfoot
% \hline
% \endlastfoot
% Based on the given scenario, it seems the incident is related to Scenario ID 5  from the table,\\ which corresponds to an "Overturned Truck" incident where a truck has stalled on the entry ramp.\\
% The identified mitigation actions to be taken could include:\\
% - Partial closure of the ramp to ensure the safety of both the workers and the traffic merging into the highway.\\
% - Implement a temporary speed limit reduction in the area to maintain a safer, slower traffic flow.\\
% - Deploy a tow truck to remove the stalled truck and a cleanup crew if any debris is present.\\
% The required equipment or technology for these actions might be:\\
% - Warning or detour signs to guide vehicles safely around the incident.\\
% - A tow truck to remove the stalled vehicle.\\
% - A cleanup crew for cleaning up any debris and ensuring the ramp is safe for vehicles after the stalled truck is removed.\vspace{0.3cm}\\
% \end{longtable}
% \end{center}

\begin{table}[h!]
\centering
\caption{ChatGPT's Incident Response Plan for Accident ID A-2760450}
\label{tab:incident_response_plan_gpt}
\begin{tabular}{|p{14cm}|}
\hline
Based on the given scenario, it seems the incident is related to Scenario ID 5  from the table,\\ which corresponds to an "Overturned Truck" incident where a truck has stalled on the entry ramp.\\
The identified mitigation actions to be taken could include:\\
- Partial closure of the ramp to ensure the safety of both the workers and the traffic merging into the highway.\\
- Implement a temporary speed limit reduction in the area to maintain a safer, slower traffic flow.\\
- Deploy a tow truck to remove the stalled truck and a cleanup crew if any debris is present.\\
The required equipment or technology for these actions might be:\\
- Warning or detour signs to guide vehicles safely around the incident.\\
- A tow truck to remove the stalled vehicle.\\
- A cleanup crew for cleaning up any debris and ensuring the ramp is safe for vehicles after the stalled truck is removed.\vspace{0.3cm}\\
\hline
\end{tabular}
\end{table}

\begin{table}[h!]
\centering
\caption{Microsoft Copilot's Incident Response Plan for Accident ID A-2760450}
\label{tab:incident_response_plan_co}
\begin{tabular}{|p{14cm}|}
\hline
Based on the provided Traffic Incident Mitigation Strategies table and the specific accident information, the incident seems to be most similar to Scenario ID 5 - Overturned Truck. Here’s the suggested response:\\
Incident Type: Overturned Truck (Stalled Truck) Severity: Moderate Location: Highway On-ramp/Off-ramp (Entry ramp to I-81 Southbound from 7th North St)\\
Action:\\
Implement a partial ramp closure to ensure the safety of other road users and facilitate the removal of the stalled truck.\\
Reduce the speed limit in the area to prevent further incidents due to sudden changes in traffic flow.\\
Deploy a tow truck and cleanup crew to remove the stalled truck and any debris that may be obstructing the ramp.\\
Equipment/Technology Required:\\
Tow Trucks to remove the stalled truck.\\
Cleanup Crew to clear any debris or hazards resulting from the incident.\\
Speed Limit Signs to enforce the reduced speed limit in the area.\\
Please note that this is a suggested response and the actual actions taken may vary based on the specific circumstances of the incident and the judgement of the incident response team on the ground. It’s always important to prioritize the safety of all road users and the quick clearance of the incident to restore normal traffic flow.\\
\hline
\end{tabular}
\end{table}

\begin{table}[h!]
\centering
\caption{Google Gemini's Incident Response Plan for Accident ID A-2760450}
\label{tab:incident_response_plan_ge}
\begin{tabular}{|p{14cm}|}
\hline
Based on the Traffic Incident Mitigation Strategies table (Table 1), the following actions are recommended for the specific accident described (Accident ID: A-2760450): \\

Based on the Traffic Incident Mitigation Strategies table (Table 1), the following actions are recommended for the specific accident described (Accident ID: A-2760450):\\
Action\\
Deploy Incident Response Vehicle\\
Temporary lane closure on the entry ramp to I-81 Southbound from 7th North St.\\
Use Variable Message Signs (VMS) to warn drivers about the stalled truck and lane closure.\\
Equipment/Technology Required\\
Service Patrol Vehicle\\
Variable Message Signs (VMS)\\
Traffic Cones or Barriers\\
Additional Information\\
The accident severity is likely low based on the description of a stalled truck.\\
Motorists should be advised to use alternate routes if possible.\\
Monitor weather conditions and adjust response as needed.\\
\hline
\end{tabular}
\end{table}

% \begin{center}
% \begin{table}{|p{14cm}|}
% \hline
% \caption{Google Gemini's Incident Response Plan for Accident ID A-2760450}\\
% \hline
% \endfirsthead
% \multicolumn{1}{c}
% {{\tablename\ \thetable{} -- continued from previous page}} \\
% \hline 
% \endhead
% \hline \multicolumn{1}{r}{{Continued on next page}} \\ 
% \endfoot
% \hline
% \endlastfoot
% Based on the Traffic Incident Mitigation Strategies table (Table 1), the following actions are recommended for the specific accident described (Accident ID: A-2760450):\\
% Action\\
% Deploy Incident Response Vehicle\\
% Temporary lane closure on the entry ramp to I-81 Southbound from 7th North St.\\
% Use Variable Message Signs (VMS) to warn drivers about the stalled truck and lane closure.\\
% Equipment/Technology Required\\
% Service Patrol Vehicle\\
% Variable Message Signs (VMS)\\
% Traffic Cones or Barriers\\
% Additional Information\\
% The accident severity is likely low based on the description of a stalled truck.\\
% Motorists should be advised to use alternate routes if possible.\\
% Monitor weather conditions and adjust response as needed.\\
% \end{table}
% \end{center}

In this study, we perform a comparative analysis of the response strategies suggested by three different AI technologies—ChatGPT, Microsoft Copilot, and Google Gemini. This evaluation, while insightful, should be noted as a brief human assessment of the responses generated by these systems. There is notable consistency in the actions recommended by ChatGPT and Microsoft Copilot, both advocating for partial closure, speed limit reduction, and the deployment of tow trucks and cleanup crews (see Table \ref{table:response_strategies}). Conversely, Google Gemini's strategy includes the deployment of an Incident Response Vehicle, temporary lane closure, and the use of Variable Message Signs (VMS), indicative of a more general approach targeted at immediate safety. ChatGPT and Microsoft Copilot provide specific actions with clear implementation steps, emphasizing a structured approach to incident management. Google Gemini, while also actionable, presents more general guidelines. All three AIs generally align with predefined guidelines, though nuances exist. ChatGPT and Microsoft Copilot closely follow these guidelines, while Google Gemini shows slight divergence, particularly in its omission of detailed cleanup advice, opting instead for more generalized response actions. This qualitative assessment underscores the difference in AI-generated strategies by various LLM models and the importance of this consideration for optimal real-world implementations.

\begin{table}
\centering
\small
\begin{tabular}{|p{3cm}|p{4cm}|p{4cm}|p{4cm}|}
\hline
\textbf{Criteria}                                & \textbf{ChatGPT}                                    & \textbf{Microsoft Copilot}                          & \textbf{Google Gemini}                              \\ \hline
\textbf{Emphasis}                                & Safety and quick clearance                          & Safety, clear communication, and quick clearance   & Immediate response actions for safety               \\ \hline
\textbf{Actions}                                 & - Partial closure\newline- Speed limit reduction\newline- Tow truck and cleanup deployment & - Partial closure\newline- Speed limit reduction\newline- Tow truck and cleanup deployment & - Incident Response Vehicle deployment\newline- Temporary lane closure\newline- Use of VMS \\ \hline
\textbf{Granularity of Actions}                  & Specific actions with clear steps for implementation & Similar specificity, emphasizing safety and traffic flow & More general actions focused on immediate safety measures \\ \hline
\textbf{Equipment Recommended}                   & - Tow trucks\newline- Cleanup crew\newline- Warning/detour signs & - Tow trucks\newline- Cleanup crew\newline- Speed limit signs & - Service Patrol Vehicle\newline- VMS\newline- Traffic Cones or Barriers \\ \hline
\textbf{Divergence from Guidelines (If applicable)} & Aligns well with predefined guidelines               & Closely follows predefined guidelines               & Slight divergence; lacks mention of cleanup         \\ \hline
\textbf{Presence of Recommended Actions in Guidelines} & Fully present, well represented                    & Fully present, well aligned                         & Partially present; lacks some detailed actions      \\ \hline
\end{tabular}
\caption{Comparison of response strategies by different AI technologies.}
\label{table:response_strategies}
\end{table}

\subsection{Incident response plan comparison methodology}

During the incident management, decision-makers are frequently confronted with the need to evaluate and select from a multitude of response actions under conditions of uncertainty and time constraints. The commonly used method is Analytical Hierarchy Process (AHP) which serves as a robust multi-criteria decision-making (MCDM) tool that assists in organizing and analyzing these complex decisions. By breaking down the decision-making problem into a hierarchy of more easily comprehended sub-problems, AHP enables decision-makers to systematically evaluate the relative importance of various criteria and actions. While the Analytical Hierarchy Process (AHP) provides a structured approach for multi-criteria decision-making (MCDM), in this study, we leverage the Technique for Order Preference by Similarity to Ideal Solution (TOPSIS) \cite{LAI1994486,zavadskas2016development} method to compare incident response strategies. TOPSIS is another robust MCDM technique that ranks alternatives based on their relative closeness to an ideal solution. Unlike AHP, which involves pairwise comparisons and eigenvector calculations, TOPSIS evaluates alternatives based on their geometric distance to a positive ideal solution (PIS) and a negative ideal solution (NIS).

The steps involved in the TOPSIS method are as follows:

1. \textbf{Construct the Decision Matrix:} 
   Let \( X \) be an \( m \times n \) matrix representing \( m \) alternatives (incident response plans) and \( n \) criteria (in our case we will consider for simplcity only 2 criteria later on , e.g. Impact and Resource Engagement). Each element \( x_{ij} \) in the matrix \( X \) represents the performance score of alternative \( i \) with respect to criterion \( j \).

   \[
   \mathbf{X} = 
   \begin{bmatrix}
   x_{11} & x_{12} & \cdots & x_{1n} \\
   x_{21} & x_{22} & \cdots & x_{2n} \\
   \vdots & \vdots & \ddots & \vdots \\
   x_{m1} & x_{m2} & \cdots & x_{mn}
   \end{bmatrix}
   \]

2. \textbf{Normalize the Decision Matrix:}
The elements of the decision matrix are normalized to transform the various criteria to a comparable scale.

\[
r_{ij} = \frac{x_{ij}}{\sqrt{\sum_{i=1}^m x_{ij}^2}}
\]

The normalized decision matrix \( \mathbf{R} \) is given by:

\[
\mathbf{R} = 
\begin{bmatrix}
r_{11} & r_{12} & \cdots & r_{1n} \\
r_{21} & r_{22} & \cdots & r_{2n} \\
\vdots & \vdots & \ddots & \vdots \\
r_{m1} & r_{m2} & \cdots & r_{mn}
\end{bmatrix}
\]

3. \textbf{Determine the Weighted Normalized Decision Matrix:}
Apply weights to the normalized matrix to reflect the relative importance of each criterion. Let \( w_j \) be the weight of criterion \( j \). The weighted normalized decision matrix \( \mathbf{V} \) is calculated as follows:

\[
v_{ij} = r_{ij} \times w_j
\]

\[
\mathbf{V} = 
\begin{bmatrix}
v_{11} & v_{12} & \cdots & v_{1n} \\
v_{21} & v_{22} & \cdots & v_{2n} \\
\vdots & \vdots & \ddots & \vdots \\
v_{m1} & v_{m2} & \cdots & v_{mn}
\end{bmatrix}
\]

4. \textbf{Identify the Positive Ideal Solution (PIS) and Negative Ideal Solution (NIS):}

The PIS \( A^+ \) is a vector that contains the maximum value for each benefit criterion and the minimum value for each cost criterion:

\[
A^+ = \{ (\max_{i} v_{ij} \mid j \in J), (\min_{i} v_{ij} \mid j \in J' ) \}, \quad j = 1, 2, \ldots, n
\]

The NIS \( A^- \) is a vector that contains the minimum value for each benefit criterion and the maximum value for each cost criterion:

\[
A^- = \{ (\min_{i} v_{ij} \mid j \in J), (\max_{i} v_{ij} \mid j \in J' ) \}, \quad j = 1, 2, \ldots, n
\]

Where \( J \) are benefit criteria and \( J' \) are cost criteria.

5. \textbf{Calculate the Separation Measures:}
Compute the Euclidean distance of each alternative from the PIS (\( S_i^+ \)) and the NIS (\( S_i^- \)).

\[
S_i^+ = \sqrt{\sum_{j=1}^{n} (v_{ij} - v_j^+)^2 }, \quad S_i^- = \sqrt{\sum_{j=1}^{n} (v_{ij} - v_j^-)^2 }
\]

6. \textbf{Calculate the Relative Closeness to the Ideal Solution:}
The relative closeness \( C_i^* \) of each alternative \( A_i \) to the ideal solution is given by:

\[
C_i^* = \frac{S_i^-}{S_i^+ + S_i^-}
\]

The relative closeness \( C_i^* \) ranges from 0 to 1, with higher values indicating alternatives that are closer to the PIS and thus more preferable.

% 7. \textbf{Rank the Alternatives:}
%    The alternatives are ranked based on their relative closeness to the ideal solution. The alternative with the highest \( C_i^* \) value is considered the most effective incident response plan.

Once the weights for each action are computed, these weights are applied to evaluate different incident response plans. Each plan, represented as a binary vector of actions, is scored by summing the weights of the included actions. The total score of a plan reflects its overall effectiveness in achieving the incident management objective, considering both impact and resource engagement.

% Pairwise comparisons are carried out for each criterion, and weights are derived through the steps described above. The Analytical Hierarchy Process (AHP) is used to derive weights for various incident response actions, allowing for the evaluation and comparison of different response plans represented as binary vectors.

 The TOPSIS method is used to derive weights for various incident response actions, allowing for the evaluation and comparison of different response plans represented as binary vectors.

Given the virtual goal (necessary for the TOPSIS evaluation only) of selecting the best incident response plan, we consider actions such as "Deploy Incident Response Vehicle," "Temporary Lane Closure," etc., and criteria such as "Impact" and "Resource Engagement." The final ranking of the binary vectors representing different action plans is obtained by computing their scores based on the derived weights.

Weight binary action vectors are necessary since when using linguistic comparison metrics there is no domain knowledge involved in comparison. For example, abscence of action "Deploy IRV" can have a significant impact on efficiency of a plan while its abscence in linguistic metrics will be similar to abscence of any other words (like notifying social media). In other words, linguistic metrics may assign different weights to actions based on linguistic features in regards to impact of these actions. This is why the domain knowledge on impact and resource engagement should be incorporated into the plan comparison approach.
 
Consider the following example plans:

\begin{itemize}
  \item Plan A: \( [1, 0, 1, 1, 0, 1, 0, 1, 0, 1] \)
  \item Plan B: \( [1, 1, 0, 0, 1, 0, 1, 0, 1, 0] \)
\end{itemize}

The total score for Plan A is calculated by:
\[
\text{Score}_A = \sum_{i=1}^{n} \left( w_{\text{total}} \times a_i \right)
\]
where \( a_i \) represents the inclusion (1) or exclusion (0) of action \( i \) in Plan A.

Thus:
\[
\text{Score}_A = \left( w_1 \times 1 + w_2 \times 0 + w_3 \times 1 + \cdots + w_{10} \times 1 \right)
\]

The calculated score of a plan represents an absolute numerical representation of engaged actions as a single number. This allows for plan comparison with consideration of impact and resource engagement. Each plan for evaluation will have a score assigned by human based on incident response guidelines. It will represent a base value of each plan. Further, response plans produced by Generative AI models will be produced and compared to this value. This way we define the method and the metric for comparison of generated incident response plans. Each comparison table contains priority indexes and not actual relevance rate.% (e.g. Notify Police \& EMS instruction is not 4 times more relevant than using a social media, but but simple 4 steps more important).

\subsection{Late fusion of model responses}

The use of binary action decomposition enables the effective application of ensemble techniques to improve the accuracy of response plans. By converting model outputs into binary action vectors, we can treat each model's response as a feature vector. This allows for the application of ensembling methods, such as late fusion, where multiple models’ outputs are combined to produce a more optimal averaged solution \cite{akilan2018effect,boulahia2021early}.

Using multiple executions of the prompt we may obtain multiple binary vectors (solutions): \( \mathbf{v_1}, \mathbf{v_2}, \dots, \mathbf{v_m} \), where \( m \) is the number of responses, and each vector is of length \( n \):

\[
\mathbf{v_1} = [v_{11}, v_{12}, \dots, v_{1n}]
\]
\[
\mathbf{v_2} = [v_{21}, v_{22}, \dots, v_{2n}]
\]
\[
\vdots
\]
\[
\mathbf{v_m} = [v_{m1}, v_{m2}, \dots, v_{mn}]
\]

The element-wise average of these \( m \) vectors is given by:

\[
\mathbf{avg} = \frac{1}{m} \left( \mathbf{v_1} + \mathbf{v_2} + \dots + \mathbf{v_m} \right)
\]

This can be written as:

\[
\mathbf{avg} = \left[ \frac{v_{11} + v_{21} + \dots + v_{m1}}{m}, \frac{v_{12} + v_{22} + \dots + v_{m2}}{m}, \dots, \frac{v_{1n} + v_{2n} + \dots + v_{mn}}{m} \right]
\]

Next, apply a rounding function, \( \text{round}(x) \), to each element of the averaged vector to obtain the final binary vector \( \mathbf{b} \):

\[
\mathbf{b} = \text{round}\left( \mathbf{avg} \right)
\]

This results in:

\[
\mathbf{b} = \left[ \text{round}\left( \frac{v_{11} + v_{21} + \dots + v_{m1}}{m} \right), \text{round}\left( \frac{v_{12} + v_{22} + \dots + v_{m2}}{m} \right), \dots, \text{round}\left( \frac{v_{1n} + v_{2n} + \dots + v_{mn}}{m} \right) \right]
\]

The resulting vector \( \mathbf{b} \) is the final binary decision vector derived by averaging and rounding the outputs from the \( m \) binary vectors.

For example, let's consider binary responses from three models for a single incident:

\[
\mathbf{r_1} = [1, 0, 1, 1, 0]
\]
\[
\mathbf{r_2} = [1, 1, 0, 1, 1]
\]
\[
\mathbf{r_3} = [0, 1, 1, 0, 0]
\]

Element-wise summation:
\[
\mathbf{sum} = \mathbf{r_1} + \mathbf{r_2} + \mathbf{r_3} = [1+1+0, 0+1+1, 1+0+1, 1+1+0, 0+1+0]
\]
\[
\mathbf{sum} = [2, 2, 2, 2, 1]
\]

Step 2: Compute the average:
\[
\mathbf{avg} = \frac{1}{3} \mathbf{sum} = \left[ \frac{2}{3}, \frac{2}{3}, \frac{2}{3}, \frac{2}{3}, \frac{1}{3} \right]
\]

Step 3: Rounding to final binary response:
\[
\mathbf{final} = \text{round}(\mathbf{avg}) = [1, 1, 1, 1, 0]
\]

The final binary response is \( \mathbf{final} = [1, 1, 1, 1, 0] \), representing the averaged decision from all models.

This approach has the potential to reduce minor reasoning errors and enhance the stability of the final generated plan. Late fusion is also critical for minimizing randomness in the generated responses and ensuring the reproducibility of results. As we propose a method for quantifying solutions based on their proximity to a manual solution, the impact of different numbers of plan generations on overall model performance can also be assessed.

\section{Results}

This section introduces the application of the TOPSIS method to compare various traffic incident response strategies across varios Generative AI models. A decision matrix is constructed based on two criteria—Impact and Resource Engagement—then normalized and weighted to reflect their relative importance. Ideal and negative-ideal solutions are determined to represent the best and worst scenarios. Separation measures are calculated to evaluate each action's proximity to these solutions. The relative closeness gives us weights for comparison of response plans represented as binary vectors.

\subsection{Application of TOPSIS method to compare strategies}

In this section, we construct the decision matrix \( X \), which details the key actions for traffic incident management based on two criteria: \textbf{Impact} and \textbf{Resource Engagement}. These criteria reflect the effectiveness of the actions and the resources required to implement them. For example, actions such as "Deploy Incident Response Vehicle (IRV)" and "Full Road Closure" score high on both criteria, indicating significant impact but also high resource usage. In contrast, actions like "Activate Emergency Operations Center (EOC)" involve fewer resources but have a lower impact. This matrix will serve as the foundation for further analysis, including normalization and ranking to determine the most effective strategies.

\[
X = \begin{bmatrix}
  \text{Action} & \text{Impact} & \text{Resource Engagement} \\
  \text{Deploy IRV} & 9 & 7 \\
  \text{Temporary Lane Closure} & 8 & 6 \\
  \text{Use VMS to Warn Drivers} & 7 & 5 \\
  \text{Notify Police \& EMS} & 6 & 4 \\
  \text{Quick Clearance Policy} & 5 & 6 \\
  \text{Use VMS \& Social Media} & 4 & 3 \\
  \text{Full or Partial Lane Closures} & 3 & 5 \\
  \text{Divert Traffic to Detour Routes} & 2 & 7 \\
  \text{Activate EOC} & 1 & 8 \\
  \text{Full Road Closure} & 10 & 9 \\
\end{bmatrix}
\]

\subsubsection{Step 2: Normalization of the Decision Matrix}
The normalization step is still necessary to make values comparable on a common scale between 0 and 1.

\[
r_{ij} = \frac{x_{ij}}{\sqrt{\sum_{k=1}^{I} x_{kj}^2}}
\]

Normalized decision matrix \( R \):
\begin{table}[h!]
\centering
\small
\caption{Normalized Decision Matrix}
\label{tab:normalized_matrix}
\begin{tabular}{lrr}
\toprule
Action & Impact & Resource Engagement \\
\midrule
Deploy IRV & 0.46 & 0.35 \\
Temporary Lane Closure & 0.41 & 0.30 \\
Use VMS to Warn Drivers & 0.36 & 0.25 \\
Notify Police \& EMS & 0.31 & 0.20 \\
Quick Clearance Policy & 0.25 & 0.30 \\
Use VMS \& Social Media & 0.20 & 0.15 \\
Full or Partial Lane Closures & 0.15 & 0.25 \\
Divert Traffic to Detour Routes & 0.10 & 0.35 \\
Activate EOC & 0.05 & 0.41 \\
Full Road Closure & 0.51 & 0.46 \\
\bottomrule
\end{tabular}
\end{table}

\subsubsection{Step 3: Weighted Normalized Decision Matrix}
Weights \( w \) for each criterion:

\[
w_{\text{Impact}} = 0.7, \quad w_{\text{Resource Engagement}} = 0.3
\]

Weighted normalized decision matrix \( V \) is presented as follows (see Table \ref{tab:weighted_matrix}).

\begin{table}[h!]
\centering
\small % This will make the font size of the table smaller
\caption{Weighted Normalized Decision Matrix}
\label{tab:weighted_matrix}
\begin{tabular}{l@{\hskip 0.5in} |r@{\hskip 0.3in}| r}
\toprule
Action & Impact & Resource Engagement \\
\midrule
Deploy IRV & 0.321 & 0.106 \\
Temporary Lane Closure & 0.285 & 0.091 \\
Use VMS to Warn Drivers & 0.250 & 0.076 \\
Notify Police \& EMS & 0.214 & 0.061 \\
Quick Clearance Policy & 0.178 & 0.091 \\
Use VMS \& Social Media & 0.143 & 0.046 \\
Full or Partial Lane Closures & 0.107 & 0.076 \\
Divert Traffic to Detour Routes & 0.071 & 0.106 \\
Activate EOC & 0.036 & 0.122 \\
Full Road Closure & 0.357 & 0.137 \\
\bottomrule
\end{tabular}
\end{table}

\subsubsection{Step 4: Determine Ideal and Negative-Ideal Solutions}

Table \ref{tab:ideal_solutions} displays the positive-ideal and negative-ideal solutions based on alternative combinations of Impact and Resource Engagement. 

% In this case, the ideal solution, \( A^+ \) (Max-Min), represents the maximum value for Impact and the minimum value for Resource Engagement, highlighting a scenario with high impact and low resource usage. On the other hand, the negative-ideal solution, \( A^- \) (Min-Max), captures the minimum value for Impact and the maximum value for Resource Engagement, illustrating a scenario with low impact and high resource usage. 

% Calculate \( A^+ \) (Ideal Solution) and \( A^- \) (Negative-Ideal Solution):

% \begin{table}[h!]
% \centering
% \caption{Ideal Solutions}
% \label{tab:ideal_solutions}
% \begin{tabular}{rrr}
% \toprule
% Optima & Impact & Resource Engagement \\
% \midrule
% A+ (Max-Min) & 0.356753 & 0.045573 \\
% A- (Min-Max) & 0.035675 & 0.136720 \\
% \bottomrule
% \end{tabular}
% \end{table}

\begin{table}[h!]
\centering
\caption{Ideal Solutions}
\label{tab:ideal_solutions}
\begin{tabular}{|r|r|r|}
\toprule
Optima & Impact & Resource Engagement \\
\midrule
A+  & 0.356753 & 0.136720 \\
A-  & 0.035675 & 0.045573 \\
\bottomrule
\end{tabular}
\end{table}

% \begin{table}[h!]
% \centering
% \caption{Ideal Solutions}
% \label{tab:ideal_solutions}
% \begin{tabular}{rr}
% \toprule
% Impact & Resource Engagement \\
% \midrule
% 0.356753 & 0.136720 \\
% 0.035675 & 0.045573 \\
% \bottomrule
% \end{tabular}
% \end{table}

\subsubsection{Step 5: Compute Separation Measures}

These equations compute the Euclidean distance between each alternative and the ideal or negative-ideal solutions by summing the squared differences between each criterion's weighted normalized value and the ideal/negative-ideal values (see Table \ref{tab:separation_measures}).

% Calculate \( S_i^+ \) and \( S_i^- \):
% \[
% S_i^+ = \sqrt{\sum_{j=1}^{2} (v_{ij} - A^+_j)^2}
% \]
% \[
% S_i^- = \sqrt{\sum_{j=1}^{2} (v_{ij} - A^-_j)^2}
% \]

% Separation measures:

\begin{table}[h!]
\centering
\caption{Separation Measures \( S_i^+ \) and \( S_i^- \) for Each Action}
\label{tab:separation_measures}
\begin{tabular}{|c|c|c|}
\hline
\textbf{Action} & \( S_i^+ \) & \( S_i^- \) \\
\hline
Deploy IRV & 0.048 & 0.177 \\
Temporary Lane Closure & 0.125 & 0.172 \\
Use VMS to Warn Drivers & 0.202 & 0.154 \\
Notify Police \& EMS & 0.278 & 0.132 \\
Quick Clearance Policy & 0.188 & 0.110 \\
Use VMS \& Social Media & 0.263 & 0.088 \\
Full or Partial Lane Closures & 0.340 & 0.066 \\
Divert Traffic to Detour Routes & 0.416 & 0.044 \\
Activate EOC & 0.493 & 0.022 \\
Full Road Closure & 0.000 & 0.221 \\
\hline
\end{tabular}
\end{table}

\subsubsection{Step 6: Calculating Relative Closeness to the Ideal Solution}
By computing the relative closeness \( C_i^* \) for each binary vector representation of incident response plans, we effectively derive a numerical score that indicates how well each action balances the criteria of Impact and Resource Engagement. These obtained values represent action weights when comparing binary vectors of incident response plans (see Table \ref{tab:topsis_results}).

\begin{table}[h!]
\centering
\begin{tabular}{l|r|r|r|r}
\toprule
Action & Impact & Resource Engagement & Topsis Score & Rank \\
\midrule
Deploy IRV & 9 & 7 & 0.861632 & 2 \\
Temporary Lane Closure & 8 & 6 & 0.749898 & 3 \\
Use VMS to Warn Drivers & 7 & 5 & 0.637243 & 4 \\
Notify Police \& EMS & 6 & 4 & 0.525487 & 5 \\
Quick Clearance Policy & 5 & 6 & 0.448632 & 6 \\
Use VMS \& Social Media & 4 & 3 & 0.315083 & 7 \\
Full or Partial Lane Closures & 3 & 5 & 0.231794 & 8 \\
Divert Traffic to Detour Routes & 2 & 7 & 0.197111 & 9 \\
Activate EOC & 1 & 8 & 0.191135 & 10 \\
Full Road Closure & 10 & 9 & 1.000000 & 1 \\
\bottomrule
\end{tabular}
\caption{TOPSIS Results}
\label{tab:topsis_results}
\end{table}

\subsection{Example evaluation}

% Table \ref{tab:incident_1} presents an example comparison of different response actions for an incident. Each action in the response plan is assigned a weight, reflecting its importance. The table evaluates four plans: the Manual Solution, Model1, Model2, and Model3. Each plan's effectiveness in implementing various actions is scored as either 0 (action not taken) or 1 (action taken). The total score for each plan is calculated by summing the product of the weights and the respective scores. The Manual Solution, with a total score of 6.745, represents the optimal set of actions. The obtained value allows to calculate difference in effectivenes of response plans.

% The table showcases the effectiveness of four response plans (Manual Solution, Model1, Model2, Model3) for a specific incident. By weighting different actions based on importance and scoring each plan's implementation (0 or 1), a clearer picture of their effectiveness emerges. Higher scores indicate better response plans. Here's a breakdown:

Table \ref{tab:incident_1} compares the effectiveness of four different response plans: the Manual Solution, Model1, Model2, and Model3, in addressing a specific incident. Each action in the response plan is assigned a weight based on its importance, and each plan is scored based on whether the action was implemented (1) or not (0). The total score for each plan is calculated as the weighted sum of actions taken. 
% The Manual Solution, with the highest score of 6.745, is considered the most effective, with other models showing varying levels of response quality.

\begin{description}
    \item[Manual Solution:] Achieving the highest score (6.745), it serves as the benchmark, correctly identifying and executing all crucial actions for an optimal response.
    \item[Model2:] Ranking second with a score of 3.401, it demonstrates a good grasp of important actions, particularly regarding lane closures and traffic management. However, it misses key elements like utilizing social media and activating the EOC.
    \item[Model1:] Scoring 2.429, it shows a less effective response compared to Model2. While correctly identifying actions like deploying response vehicles and notifying authorities, it fails to implement critical traffic management strategies such as lane closures and quick clearance policies.
    \item[Model3:] With the lowest score of 2.097, it highlights inconsistencies in its response. While recognizing the need for response vehicles and lane closures, it neglects important aspects like using VMS for driver warnings and implementing quick clearance.
\end{description}

% \section*{Example of Differences in Absolute Model Score and Percentage Deviation}

% Comparing the Manual Solution and Model1 reveals the impact of score differences:

% \begin{align*}
%     \text{Difference in Total Score:} & \quad 6.745 \, (\text{Manual Solution}) - 2.429 \, (\text{Model1}) = 4.316 \\
%     \text{Percentage Deviation:} & \quad \left( \frac{4.316}{6.745} \right) \times 100\% \approx 64\%
% \end{align*}

This significant difference of 4.316, representing a 64\% deviation from the ideal score, highlights that Model1 misses several crucial actions. For instance, failing to activate the "Quick Clearance Policy" (weight = 0.369) and "Temporary Lane Closure" (weight = 0.579) significantly contributes to the score difference, emphasizing their importance in effective responses.

\begin{table}[h!]
\centering
\begin{tabular}{|c|c|c|c|c|c|}
\hline
\textbf{Action} & \textbf{Weight} & \textbf{Manual Solution} & \textbf{Model 1} & \textbf{Model 2} & \textbf{Model 3} \\ \hline
Deploy Incident Response Vehicle     & 0.787 & 1 & 1 & 1 & 1 \\ \hline
Temporary Lane Closure               & 0.579 & 1 & 0 & 1 & 1 \\ \hline
Use VMS to Warn Drivers              & 0.432 & 1 & 1 & 1 & 0 \\ \hline
Notify Police \& EMS                 & 0.322 & 1 & 1 & 1 & 1 \\ \hline
Quick Clearance Policy               & 0.369 & 1 & 0 & 1 & 0 \\ \hline
Use VMS \& Social Media              & 0.251 & 1 & 1 & 0 & 1 \\ \hline
Full or Partial Lane Closures        & 0.162 & 1 & 0 & 1 & 1 \\ \hline
Divert Traffic to Detour Routes      & 0.095 & 1 & 1 & 1 & 0 \\ \hline
Activate EOC                         & 0.042 & 1 & 1 & 0 & 1 \\ \hline
Full Road Closure                    & 1.000 & 0 & 0 & 0 & 0 \\ \hline
\textbf{Total Score} & & \textbf{6.745} & \textbf{2.429} & \textbf{3.401} & \textbf{2.097} \\ \hline
\textbf{Rank} & & \textbf{1} & \textbf{3} & \textbf{1} & \textbf{2} \\ \hline
\end{tabular}
\caption{Example comparison of response plans for an incident}
\label{tab:incident_1}
\end{table}

\subsection{Initial evaluation}

The Table \ref{tab:compfull_transposed_reduced} presents an example of the binary arrays generated by different language models for various accident IDs, as well as an "Manual Solution" for reference. The binary arrays indicate whether a specific action (each digit representing an action) was triggered (1) or not (0) by each model for the given accidents. The comparison and observations were derived from larger sample of 10 incident plans.

Gemini 1.5 Pro Model: Tends to include too many actions in response to the accidents. For many accident IDs, Gemini 1.5 Pro consistently triggers all actions (indicated by arrays filled mostly with 1s), which suggests an over-reaction or lack of discrimination between necessary and unnecessary actions. 2. GPT-4 Model: Responses appear more nuanced and selective. The binary arrays generated by GPT-4 show greater alignment with the "Manual Solution" relative to other models. For many accidents, the actions taken by GPT-4 closely match those recommended by the ideal response plan, indicating a higher level of accuracy and potentially more optimal decision-making.

GPT-4 Original (GPT-4o) and ChatGPT 3.5 also show a reasonable degree of alignment with the Manual Solution but are less consistent compared to GPT-4.
Gemini 1.5 Flash displays a tendency to over-respond, albeit not as extensively as Gemini 1.5 Pro. The variations across models suggest differing degrees of precision and over-inclusivity, highlighting the importance of selecting the appropriate language model for generating optimal response plans in accident scenarios.

Overall, the table illustrates that while the Gemini 1.5 Pro model often triggers excessive actions, GPT-4's responses are well-calibrated and closely match the ideal response strategies for the given accidents. This makes GPT-4 a potentially more reliable choice for generating effective response plans.

\begin{table}[h]
\label{tab:compfull_transposed_reduced}
\scriptsize
\centering
\begin{tabular}{|c|c|c|c|c|c|}
\hline
\textbf{} & \textbf{A-4259643} & \textbf{A-5128843} & \textbf{A-4227983} & \textbf{A-4888575} & \textbf{A-5968770} \\
\hline
\textbf{GPT-4} & [0 1 1 0 0 0 0 0 1 0] & [1 1 1 1 1 1 1 1 0 1] & [1 1 1 1 1 1 1 0 0 0] & [0 1 1 1 1 1 1 1 1 1] & [0 1 1 1 1 1 0 0 0 0] \\
\hline
\textbf{GPT-4o} & [0 0 1 1 0 1 1 1 1 1] & [0 0 1 1 1 1 1 1 0 1] & [0 0 1 0 0 1 1 1 1 1] & [0 0 1 1 0 1 1 1 1 1] & [0 0 1 1 0 0 1 1 0 1] \\
\hline
\textbf{Gemini 1.5 Flash} & [1 0 1 1 1 1 1 1 1 1] & [1 1 1 1 1 1 1 1 0 1] & [1 1 1 1 0 1 1 1 0 1] & [1 1 1 1 0 1 1 1 1 1] & [1 1 1 1 1 1 1 1 0 1] \\
\hline
\textbf{Gemini 1.5 Pro} & [1 1 1 1 1 1 1 1 1 1] & [1 1 1 1 1 1 1 1 1 0] & [1 1 1 1 1 1 1 1 1 0] & [1 1 1 1 1 1 1 1 0 1] & [1 1 1 1 1 1 0 1 0 0] \\
\hline
\textbf{ChatGPT 3.5} & [1 1 1 1 1 1 1 1 1 1] & [1 1 1 1 1 0 0 0 0 1] & [1 1 1 1 1 1 1 1 1 1] & [1 1 1 1 1 0 0 0 0 0] & [1 1 1 0 1 0 0 1 0 0] \\
\hline
\textbf{Manual Solution} & [0 0 1 1 0 1 1 1 1 1] & [1 1 1 1 1 1 1 1 0 1] & [1 1 1 1 0 1 1 1 0 0] & [0 1 1 1 0 1 1 1 1 1] & [0 1 1 1 1 1 0 1 0 0] \\
\hline
\end{tabular}
\caption{Comparison of binary arrays generated by different language models for 5 different accidents, along with the Manual Solution.}
\end{table}

\subsection{Response alignment with manual solutions}

The Table \ref{tab:comp} presents an analysis of how closely various models' response plans align with an Manual Solution for several specific incident IDs. The numerical values in the table represent the plan scores for each model and accident description, indicating the deviation of each model's response from the ideal solution - the lower the difference score, the closer the model's response is to the ideal.

The difference scores for Gemini 1.5 Pro are notably high across most incident IDs. This model tends to include too many actions. The average difference score of 1.52 further suggests that Gemini 1.5 Pro consistently over-responds, triggering excessive actions beyond what is necessary according to the Manual Solution.

GPT-4 shows very low difference from Manual Solution for most incident IDs, with an average difference of just 0.68. This indicates that GPT-4’s responses are well-calibrated and more closely aligned with manual response plans. Its low difference scores signify a high level of accuracy in selecting only the necessary actions.

GPT-4 Omni (GPT-4o) and ChatGPT 3.5 models exhibit reasonable alignment with the Manual Solution, with average difference scores of 1.16 for GPT-4o. However, they are not as consistent as GPT-4. This model performs better than Gemini 1.5 Pro but still show some deviations from the manual solution.

The difference scores for Gemini 1.5 Flash are also relatively high, with an average score of 1.15 with a tendency to over-respond, similar to Gemini 1.5 Pro, though not as excessively.

Overall, Gemini 1.5 Pro is less discriminative and tends to over-trigger actions, making it less reliable for generating optimal response plans. GPT-4 stands out for its high accuracy and precision, closely matching the human-made response plans and thereby presenting as a more effective model for decision-making.
Other Models (GPT-4o, ChatGPT 3.5, Gemini 1.5 Flash) fall somewhere in-between, showing varying degrees of precision but generally less optimal compared to GPT-4.

\begin{table}
\begin{center}
\caption{Model comparison results with average differences}
\label{tab:comp}
\begin{tabular}{|l|l|l|l|l|l|}
\toprule
 & GPT-4 & GPT-4o & Gemini 1.5 Flash & Gemini 1.5 Pro & Manual solution \\
\midrule
A-4259643 & 1.81 & 2.77 & 4.30 & 5.16 & 2.77 \\
A-5128843 & 4.96 & 3.10 & 4.96 & 4.97 & 4.96 \\
A-4227983 & 4.54 & 2.13 & 4.44 & 4.97 & 4.24 \\
A-4888575 & 4.16 & 2.77 & 4.63 & 4.96 & 3.63 \\
A-5968770 & 3.22 & 2.13 & 4.96 & 4.45 & 3.45 \\
A-6137133 & 0.75 & 1.57 & 4.96 & 4.77 & 1.93 \\
A-4428281 & 2.77 & 3.63 & 3.70 & 4.44 & 2.70 \\
A-4732415 & 0.75 & 3.57 & 4.22 & 4.63 & 2.84 \\
A-6060568 & 4.63 & 1.61 & 4.63 & 4.96 & 4.63 \\
A-3996497 & 0.75 & 2.57 & 4.01 & 5.16 & 2.15 \\
\midrule
Average Difference & 0.68 & 1.16 & 1.15 & 1.52 & 0.00 \\
\bottomrule
\end{tabular}
\end{center}
\end{table}

For detailed list of generated responses as well as considered accidents, please refer to the code: \url{https://github.com/Future-Mobility-Lab/TrafficIncidentGPT}.

% \subsection{Observed response defects}

\subsection{Analysis of Hallucinations in LLMs During Plan Generation}

In the evaluation of large language models (LLMs), several types of hallucinations were observed across multiple model generations:

\begin{itemize}
    \item \textbf{Omission and invention of non-existing actions:} The models GPT-3.5-Turbo and GPT4 generated a binary set of actions, either reducing the expected 10 actions (e.g. to 6) or inventing additional actions (e.g. GPT4 produced an array of 24 actions instead of 10), which highlights its tendency to hallucinate and invent actions that were not specified on some ocassions.
    
    \item \textbf{Overreaction:} The Gemini 1.5 Pro model consistently overreacted by including all available actions in its response, indicating impaired reasoning abilities. It lacked the discernment to filter unnecessary actions, which may suggest excessive conservatism in action selection.

    \item \textbf{Indirect Response:} Instead of generating a direct binary array, the GPT-3.5-Turbo model produced Python scripts for generating the binary action array. The model may use alternative representation format unsuited to the prompt.

    \item \textbf{Formatting Variability:} All models produced responses in different markup languages: Markdown, LaTeX, Python code and plain text—demonstrating inconsistencies in output format.

    \item \textbf{Placement of Reasoning:} In all instances, models placed reasoning for their action choices mixed with the binary array, rather than delivering a final response in the form of binary array, leading to potential difficulty of extracting the required array.

    \item \textbf{Randomness in responses:} All the models tend to generate different response every time the prompt is evaluated. These variations may contain any of the mentioned previously issues.

\end{itemize}

Overall, the variability in responses, response format and structure necessitates additional prompting and processing steps to accurately extract the binary action array. The prompt may also need the re-evaluation if the response does not contain the necessary answer.

\section*{Conclusion}

This study demonstrates that generative AI models hold significant potential for revolutionizing traffic incident management. However, our analysis reveals notable differences in the accuracy and effectiveness of various language models when generating response plans. GPT-4 consistently outperforms other models, generating plans that closely align with human-proposed solutions. 

%For example, in a scenario involving a multi-vehicle collision during peak hours (Incident ID A-4227983), GPT-4 accurately identified the need for temporary lane closures and lack of need of alternate actions. This contrasted with Gemini 1.5 Pro, which, despite recognizing the need for response vehicles, neglected critical actions like VMS deployment, leading to a less effective response strategy. 

We acknowledge that deploying AI in traffic management presents challenges such as potential data and guideline bias (since inherently data-driven approach) and ethical considerations surrounding automated decision-making.

Despite these challenges, the potential benefits of AI-driven traffic management cannot be overstated. The ability of these models to swiftly generate complex, adaptable, and transparent (via chain-of-thought process) response plans has potential for far-reaching implications for improving safety, mitigating incident impact on traffic flow and reducing the economic and environmental impact of road incidents. This research serves as a crucial step toward realizing the full potential of AI in automated management of incidents.

% Urban traffic management presents a significant challenge due to the continuous increase in road incidents and resulting traffic congestion. Such issues are compounded by the inherent complexity of urban traffic networks, thus necessitating novel, effective solutions. This research introduces IncidentResponseGPT, an innovative Generative AI platform designed to enhance traffic incident response plans in real-time. It accomplishes this by employing a generative language model like ChatGPT and incorporating real-time traffic incident reports and region-specific operational guidelines.

% In summary, the comprehensive analysis of the different language models' performance in generating response plans for various accident scenarios highlights distinct differences in their decision-making accuracy and effectiveness. The data clearly shows that GPT-4 is the most reliable and optimal choice, consistently producing response plans that closely align with the ideal solutions. Its low average difference score (from ideal solution) underscores its ability to selectively trigger necessary actions without overreacting. Conversely, the Gemini 1.5 Pro model tends to over-respond significantly, with high difference scores indicating a lack of discriminative logic in its decision-making process.

% In the future research, the produced response plans are expected to have a formal structure to facilitate validation using traffic simulation. Also, each response will include references to specific parts of guidelines and Chain-of-Tought process.

In the future research, the produced response plans are expected to have a formalized structure to facilitate validation of each plan using traffic simulation. 

% The application of IncidentResponseGPT in a realistic traffic management scenario could revolutionize how traffic incidents are managed. By swiftly generating informed and adaptable response plans, IncidentResponseGPT has the potential to greatly enhance traffic safety, reduce congestion and consequent economic and environmental impacts.

% This research serves as the foundation for the deployment of IncidentResponseGPT in real-world scenarios, providing a key example of the practical application of AI systems that align with complex, real-world objectives and human-centric values, such as improving the efficiencies of traffic management and safety.

% The complexity of traffic incident response and the potential benefits of employing generative AI makes this an exciting area of research. Successful implementation of IncidentResponseGPT could open the door to future applications of AI in other complex situations that would benefit from real-time, adaptable, and transparent decision-making.

\bibliography{bib}

\end{document}